\documentclass[letterpaper]{article} 
\usepackage{aaai2026}  
\usepackage{times}  
\usepackage{helvet}  
\usepackage{courier}  
\usepackage[hyphens]{url}  
\usepackage{graphicx} 
\usepackage{natbib}  
\usepackage{caption} 
\usepackage{algorithm}
\usepackage{algorithmic}
\usepackage{amsmath} 
\usepackage{amssymb}
\usepackage{multirow}
\usepackage{subcaption}
\usepackage{pdfpages}
\usepackage{xcolor}

\usepackage{newfloat}
\usepackage{listings}
\DeclareCaptionStyle{ruled}{labelfont=normalfont,labelsep=colon,strut=off} 
\floatstyle{ruled}
\newfloat{listing}{tb}{lst}{}
\floatname{listing}{Listing}
\nocopyright 

\title{3D-CDRGP: Towards Cross-Device Robotic Grasping Policy in 3D Open World}

\author{Weiguang Zhao\textsuperscript{1,2,3*}, Chenru Jiang\textsuperscript{3}\thanks{Equal contribution}\, Chengrui Zhang\textsuperscript{1,2}, Jie Sun\textsuperscript{2}, Yuyao Yan\textsuperscript{2}, Rui.Zhang\textsuperscript{2}\thanks{Corresponding authors},  Kaizhu Huang\textsuperscript{3$\dagger$}\\
\textsuperscript{1}University of Liverpool $\quad$  \textsuperscript{2}Xi'an Jiaotong-Liverpool University $\quad$ \textsuperscript{3}Duke Kunshan University   \\
{\tt\small \{weiguang.zhao, aruix\}@liverpool.ac.uk \quad \tt\small cj262@duke.edu}  \\
{\tt\small \{rui.zhang02, Jie.Sun, Yuyao.Yan\}@xjtlu.edu.cn \quad \tt\small kaizhu.huang@dukekunshan.edu.cn}
}

\usepackage{bibentry}

\begin{document}

\maketitle

\begin{abstract}
Given the diversity of devices and the product upgrades, cross-device research has become an urgent issue that needs to be tackled. To this end, we pioneer in probing the cross-device (cameras \& robotics) grasping policy in the 3D open world. Specifically, we construct two real-world grasping setups, employing robotic arms and cameras from completely different manufacturers.  To minimize domain differences in point clouds from diverse cameras, we adopt clustering methods to generate 3D object proposals. However, existing clustering methods are limited to closed-set scenarios, which confines the robotic graspable object categories and ossifies the deployment scenarios. To extend these methods to open-world settings, we introduce the SSGC-Seg module that enables category-agnostic 3D object detection. The proposed module transforms the original multi-class semantic information into binary semantic cues—foreground and background by analyzing the SoftMax value of each point, and then clusters the foreground points based on geometric information to form initial object proposals. Furthermore, ScoreNet$^\ddagger$ is designed to score each detection result, and the robotic arm prioritizes grasping the object with the highest confidence score. Experiments on two different types of setups highlight the effectiveness and robustness of our policy for cross-device robotics grasping research. Our code is provided in the supplementary and will be released upon acceptance.
\end{abstract}

\section{Introduction}
Robotic grasping seeks to manipulate objects via grippers or dexterous hands, enabling fundamental operations such as object pick-up and placement. It plays a crucial role in various manufacturing and automation applications~\cite{d2023ros,luo2024omnigrasp,zheng2024gaussiangrasper,tang2025foundationgrasp}. Driven by the rich geometric cues inherent in 3D point clouds, 3D-vision-based robotic grasping has garnered increasing attention~\cite{wang2023dexgraspnet,zhuang2023instance,zhuang2024attentionvote,zhong2025dexgrasp}. However, these approaches are typically confined to fixed scenarios, constricting both the vision sensor and robotic arm to specific versions while limiting the graspable objects to a predetermined set. With the accelerating cadence of hardware upgrades, this issue markedly exacerbates the burden of grasp system migration. 

\begin{figure}[h]
  \centering 
  \includegraphics[width=0.47\textwidth]{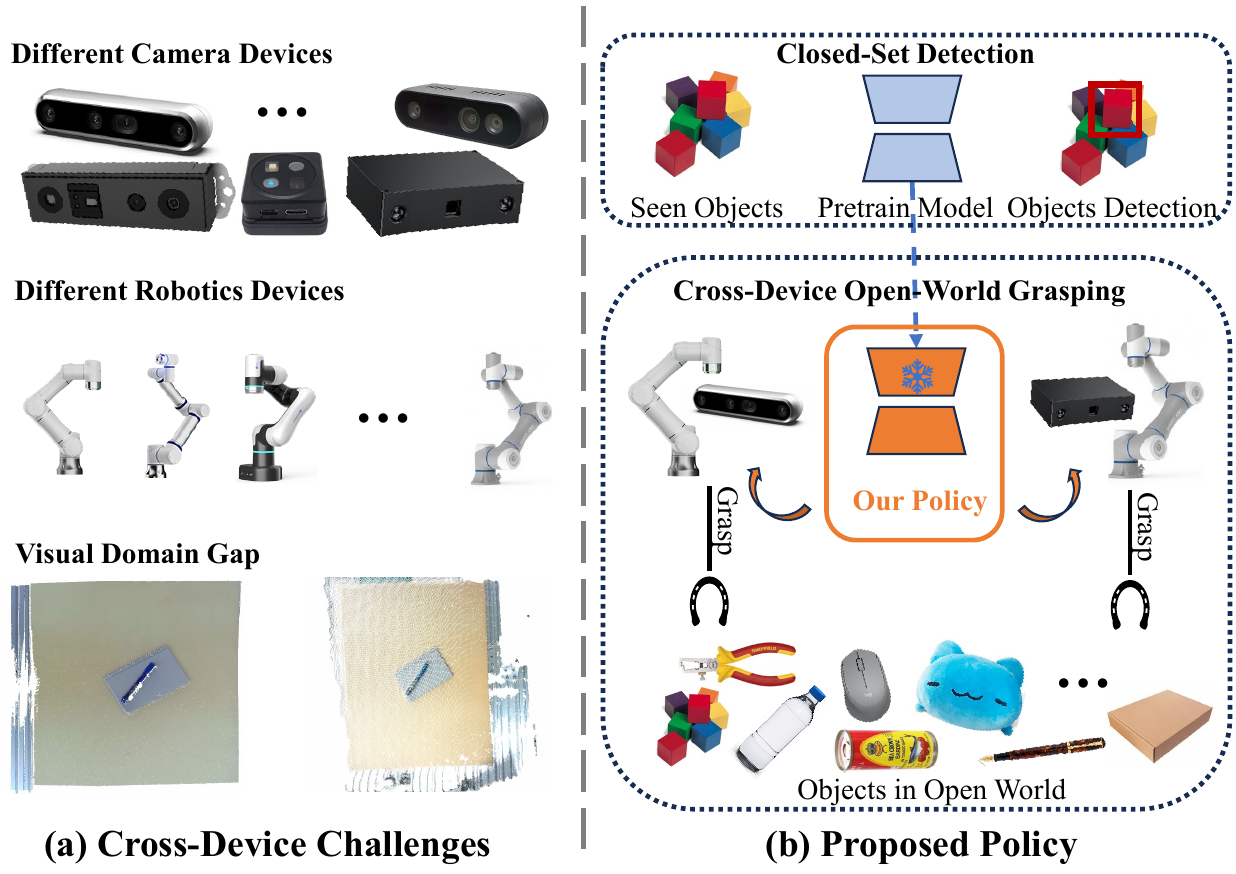}
\caption{(a) Cross-Device Challenges. The first and second rows display vision cameras and robotic arms from different manufacturers and versions; the third row presents 3D point clouds captured by an AiNSTEC PRO and a RealSense D455, respectively. (b) Proposed Policy. Our policy transfers a closed-set 3D object detector to two distinct robotic-arm grasping setups to perform class-agnostic grasping in the 3D open world. The snowflake  icon signifies that our policy operates without any additional training on the pre-trained model.}
  \label{fig:intro}
\end{figure}

As depicted in Fig.~\ref{fig:intro}~(a), we showcase vision cameras and robotic arms from distinct manufacturers and versions in the first and second rows, respectively. Owing to the rigid mechanical architecture of robotic arms, cross-device transfer is readily achieved by revising kinematic trajectories and substituting software development kit (SDK) interfaces~\cite{ekrem2023trajectory,liu2023target,Ze2024DP3}.  In contrast, the domain gap in 3D point clouds arising from visual sensor heterogeneity constitutes the critical challenge that warrants immediate resolution. Since point clouds produced by different visual devices exhibit a pronounced domain gap, manifesting in disparities of density, height, and color, as illustrated in the third row of Fig.~\ref{fig:intro}~(a). Furthermore, closed-set training rigidly confines the graspable object categories and ossifies the deployment scenarios.

To this end,  we pioneer the establishment of a 3D open-world grasping policy aimed at mitigating the domain discrepancy in 3D point-cloud inputs induced by visual-device heterogeneity, and expanding the spectrum of graspable object categories beyond closed-set constraints. On one hand, we propose employing the clustering-based 3D object detection methods~\cite{jiang2020pointgroup,chen2021hierarchical,vu2022softgroup,pbnet} to address the cross-visual-device issue, given its insensitivity to point cloud density and deep learning features. Considering that existing clustering-based 3D object detection algorithms are typically pre-trained on closed-set datasets, we propose a SSGC-Seg module to modify the semantic and point-offset dual branches of the original network. Specifically, we observe that the semantic branch consistently assigns the highest SoftMax value to the background class, enabling the identification of background points, while the remaining points are regarded as foreground points. Furthermore, we achieve the foreground object proposals by applying the geometry-based clustering methods~\cite {jiang2020pointgroup,pbnet} to these foreground points, enabling our policy's 3D open-world category-agnostic grasping capability.

On the other hand, the existing ScoreNet~\cite{jiang2020pointgroup,liang2021instance,pbnet} relies entirely on deep learning features derived from point clouds to score object proposal confidence. However, the significant domain gap in point clouds greatly affects these deep learning features, limiting ScoreNet's effectiveness in real-world grasping scenarios. To alleviate this issue, we develop an improved scoring mechanism, termed ScoreNet$^\ddagger$, which revises the object proposal scoring by incorporating the number of points within the corresponding proposal and its height. This modification helps eliminate false detections and optimizes the grasping sequence.

To comprehensively evaluate our policy, we not only validate its effective performance on segmentation datasets based on existing clustering-based 3D detection methods~\cite{jiang2020pointgroup,pbnet}, but also establish two cross-device grasping systems in real-world environments to assess its performance in practical scenarios. The contributions of our work are as follows:
\begin{itemize}
    \item Our work pioneers one cross-device 3D open-world robotic grasping policy that enables plug-and-play deployment across heterogeneous vision systems and robotic platforms without retraining.
    \item Our policy introduces the SSGC-Seg module to adapt closed-set 3D detectors for open-world object proposal detection, and develops ScoreNet$^\ddagger$  to enhance detection and grasping robustness. Moreover, the proposed policy exhibits broad applicability, being compatible with most existing clustering-based 3D object detection methods.
    \item  Comprehensive evaluations on both the segmentation dataset and real-world cross-device robotic setups demonstrate the effectiveness and generalization capability of the proposed approach in class-agnostic, domain-invariant grasping scenarios.

\end{itemize}

\begin{figure*}[t]
	\centering
	\includegraphics[width=0.99\textwidth]{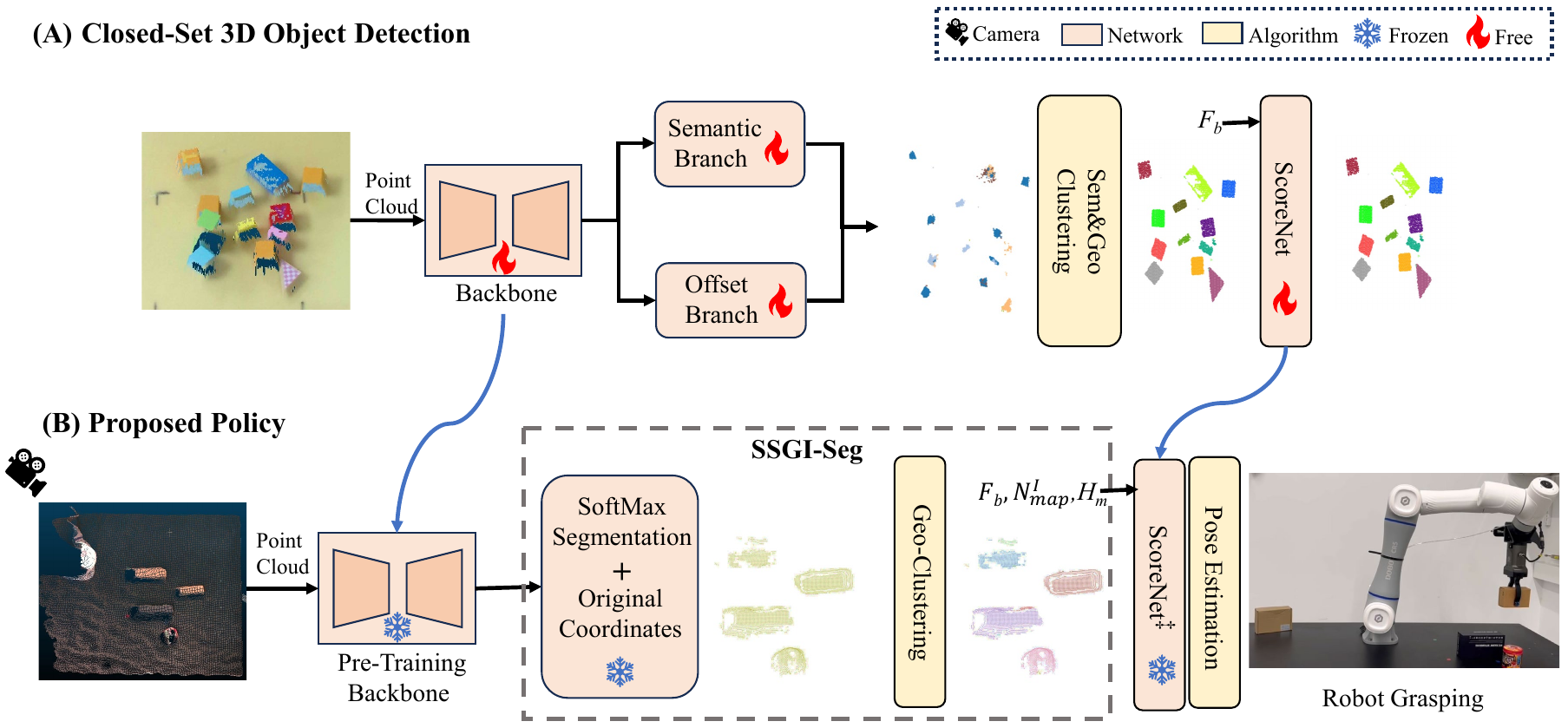} 
	\caption{Overview of Proposed Policy. (A)Closed-Set 3D Object Detection denotes the canonical pipeline for clustering-based 3D object detection~\cite{jiang2020pointgroup,pbnet}, while the (B)Proposed Policy implies the migration of the closed-set 3D object detection pipeline to the robot grasping approach across devices without additional training. $F_b$ stands for the backbone feature. $N_{map}^{I}$ is the number of points for each prediction object mask. $H_m$ denotes the mean height of the prediction object points.}
	\label{fig:pipeline}
\end{figure*}

\section{Related Work}
\subsection{Robotic Grasping with Deep Learning}
Robotic grasping aims to manipulate objects using grippers or dexterous hands for tasks such as picking up and placing down objects~\cite{fang2020graspnet,tong2024novel,fu2024light,yin2025dexteritygen}. Traditional approaches~\cite{bicchi2000robotic,AinetterF21} typically operate this task in a highly controlled way, focusing on specific or predefined objects.  With the development of deep learning, researchers have started to design various neural networks to handle grasping tasks. Recently, some studies~\cite{zhuang2023instance,kim2024vision,zhuang2024attentionvote} attempt to utilize 3D encoders to obtain point cloud representations. ISPC~\cite{zhuang2023instance} adopts the PointNet~\cite{qi2017pointnet} to extract the point cloud feature to segment the object instance. In addition, AttentionVote~\cite{zhuang2024attentionvote} develops the VoteNet~\cite{qi2019deep} with an attention mechanism to achieve higher-quality instance predictions. However, these methods are confined to closed-set scenarios, meaning they can only detect objects from predefined categories. In contrast, our method combines the 3D encoder and the proposed SSGC-Seg module to acquire rich geometric features from point clouds and enable category-agnostic open-world grasping tasks.

\subsection{Closed-Set 3D Object Detection }
3D object detection~\cite{qi2019deep,jiang2020pointgroup,zhong2022maskgroup,dong2022learning,pbnet,zhang2024safdnet} enables the localization and semantic interpretation of objects within a three-dimensional spatial context. It is extensively utilized in numerous practical applications, including autonomous driving~\cite{xu2024drivegpt4,hu2024ow3det} and robotic manipulation~\cite{duan2024aha,liu2024self}. Specifically, VoteNet~\cite{qi2019deep} presents deep hough voting, an algorithm that detects 3D objects in point clouds by predicting object centers and orientations through a voting mechanism, followed by clustering and hypothesis verification to accurately identify object instances. Additionally, Pointgroup~\cite{jiang2020pointgroup}  proposes a dual-set grouping method for 3D instance segmentation, clustering points into instances via semantic predictions and spatial relationships. Furthermore, PBNet~\cite{pbnet} designs a divide-and-conquer approach for 3D point cloud instance segmentation, utilizing point-wise binarization to efficiently separate and identify individual instances. These methods primarily focus on closed-set scenarios, whereas real-life applications involve an open world where object categories and shapes are not fixed. In this work, we pioneer the exploration of transferring closed-set 3D methods to open-world robotic arm grasping tasks, leveraging the proposed SSGC-Seg module to achieve category-agnostic 3D object detection.

\subsection{Cross-Device Transfer Learning}
In recent years, the proliferation of devices with diverse functionalities, accuracies, and models has been observed. Correspondingly, research related to cross-device deployment~\cite{bohg2014data,hirose2023exaug,cheng2024open,cheng2024navila,bjorck2025gr00t} starts to emerge, albeit in limited quantity.  ExAug~\cite{hirose2023exaug} focuses on the differences in visual devices and proposes to perform data augmentation on the training point clouds to mitigate the input differences caused by cross-device variations. More work~\cite{cheng2024open,cheng2024navila,bjorck2025gr00t} is dedicated to using multiple hardware devices to verify the scalability of the methods. These methods are often adapted based on hardware structures and do not involve research on the robustness of visual inputs. In contrast, our work paves the way to the exploration of policy to mitigate the impact of differences in 3D point cloud inputs caused by variations in visual devices.

\section{Revisiting Closed-Set 3D Object Detection Pipeline}
As shown in Fig.~\ref{fig:pipeline}~(A), we provide a concise description of the clustering-based closed-set 3D object detection pipeline~\cite{jiang2020pointgroup,chen2021hierarchical,pbnet}. The input to the network consists of $xyz$ coordinates, $rgb$ color information. This combined input is fed into a 3D UNet, which serves as the backbone. The UNet extracts features and then, through semantic and offset branches, predicts point-wise semantic labels and distance offsets from object centers. These predictions are then used for semantic\&geometry clustering, where points are first coarsely categorized by semantics, then grouped into instance proposals via geometry-driven clustering that exploits their positional or density cues. For example, PointGroup~\cite{jiang2020pointgroup} clusters points based on the distances between neighboring points, while PBNet~\cite{pbnet} performs clustering according to the density of adjacent points.
Moreover, the backbone features $F_b$ of these proposals are fed into ScoreNet to compute a confidence score for each proposal. In this regard,  objects with semantic information can be obtained based on the instance segmentation results. In this work, we make an early attempt to develop a policy that directly leverages the pretrained models of this pipeline for cross-device 3D robotic grasping applications in the open world. Moreover, the proposed policy is designed to be compatible with the majority of clustering-based 3D object detection algorithms and adaptable to a wide range of 3D vision sensors.

\section{Our Policy}

\subsection{Problem Statement}
Formally, our goal is to leverage a point cloud pre-trained object detection model trained on a limited number of seen class samples and transfer it to different types of cameras and robotic arm systems. It enables the system to grasp arbitrary objects without the need for retraining or fine-tuning,  and the grasping process does not consider the category of the objects. Specifically, given a set of rectangular building blocks $\mathcal{B}$ as the seen classes, and an object detection model $\mathcal{M}$ pre-trained on $\mathcal{B}$, the objective of this task lies in the policy to transfer $\mathcal{M}$ to a robotic arm grasping system composed of different devices without retraining or fine-tuning, thereby enabling the system to grasp arbitrary objects, including those from unseen classes. This work validates the approach using two different sets of cameras $\mathcal{CAM}_a$ and $\mathcal{CAM}_b$, and robotic arms $\mathcal{ARM}_a$ and $\mathcal{ARM}_b$.

\subsection{Overview}
As shown in Fig.~\ref{fig:pipeline}, we offer a universal 3D open-world robot grasping pipeline across devices~(cameras \& robots), which leverages the pre-training model of he closed-set 3D object detection pipeline to detect and score objects directly. Specifically, our proposed pipeline includes mainly: 1)~Data collection, 2)~SSGC-Seg, 3)~ScoreNet$^\ddagger$, and 4)~Robot grasping.  First, we feed the point clouds ($xyz$) into the backbone network to achieve the point cloud feature. Moreover, we optimize the semantic branch by analyzing the SoftMax value of each point in the closed-set 3D object detection pipeline to obtain the foreground prediction and remove the background points. Furthermore, we employ the geo-clustering~\cite{pbnet} to get the object instance and utilize the ScoreNet to assign scores to each instance, thus determining the grasping order for the robot. Finally, the object segmentation results serve as the basis for the robot's grasping posture. More details of the methods are presented in the following sections.

\subsection{Data Collection}
Our method is based on the eye-to-hand approach, with a camera fixed in a suspended position above the grasping scene. At the outset, we extract a frame of video data and convert it into point cloud format, comprising coordinate data $xyz \in \mathbb{R}^{N \times 3}$, texture data  $rgb \in \mathbb{R}^{N \times 3}$, and normal data $nl \in \mathbb{R}^{N \times 3}$, where $N$ stands for the number of points. Note that the coordinate data is a must, while texture and normal data are both optional. Moreover, the camera position and brand can be changed, and after any changes, hand-eye calibration is performed to establish the relationship between the camera coordinate system and the robot coordinate system. In this paper, we transform the camera coordinate system to the robot coordinate system, and through hand-eye calibration, we could obtain the corresponding rotation matrix  $R \in \mathbb{R}^{3 \times 3}$ and translation matrix $T \in \mathbb{R}^{3 \times 3}$.

\subsection{SSGC-Seg}

\subsubsection{SoftMax Segmentation} 
Following existing 3D segmentation and detection works~\cite{jiang2020pointgroup, he2021dyco3d, wu2022dknet, pbnet}, the reference pipeline converts the point cloud into a voxel format and utilizes MinkUnet~\cite{li2019gs3d} for sparse convolution~\cite{Graham2015Sparse3C} to obtain voxel features. Subsequently, it converts the voxels back into a point cloud to obtain features. In addition, two branches consisting of MLP layers are leveraged to predict the semantic scores $\mathbf{S} \in[0,1]^{N \times M}$ and the offset  $\boldsymbol{o}_i=\left\{o_x^i, o_y^i, o_z^i\right\}$ from the center of the object for each point, where $N$ and $M$ are the number of point and class, where $i \in\{1, \ldots, N\}$. 

Since the closed-set 3D object detection method is trained on a dataset containing specific category objects, its semantic branch cannot be applied for open-world object segmentation. It is observed that the background class still maintains relatively high softmax values; we categorize points into two classes (foreground and background) by analyzing the SoftMax value of each point. The foreground includes the objects to be grasped, while the background refers to items such as the tabletop that do not require analytical manipulation. In this regard, we can attain the binary classification score $S_b$ for the foreground and background. The binary classification score $\mathbf{S_b} \in[0,1]^{N \times 2}$ can be calculated by the SoftMax value $\mathbf{S} \in[0,1]^{N \times M}$, from the semantic branch by the following:
\begin{equation}
    \mathbf{S^i_b}= [\mathbf{S^i[1]},\ max(\mathbf{S^i[2:M]})] \quad i \in[1, N],
    \label{eq:1}
\end{equation}
where $\mathbf{S^i_b}$ and $\mathbf{S^i}$ represent the binary classification score and semantic SoftMax value for the $i_{th}$ point, respectively. Additionally, $max(\cdot)$ means the maxpooling operation and $2:M$ stands the data collection from $2_{th}$ to $M_{th}$ index. Furthermore, the binary classification prediction $\mathbf{P_b} \in~\{0, 1\}$ can be obtained by the following formula: 

\begin{equation}
    \mathbf{P^i_b}= argmax(\mathbf{S^i_b}) \quad i \in[1, N],
    \label{eq:2}
\end{equation}
where $\mathbf{P^i_b}$ means the binary classification prediction for $i_{th}$ point. Moreover, $argmax(\cdot)$ stands for the operation of taking the index of the maximum value. $\mathbf{P^i_b}=0$ implies $i_{th}$ point is the background point; otherwise, it is the foreground point.

\subsubsection{Geo-Clustering}

The clustering algorithm does not contain network parameters, enabling it to robustly adapt to cross-device 3D object detection. To our best knowledge, PBNet~\cite{pbnet} is the best clustering-based 3D instance segmentation method. Its binary clustering is employed to attain the preliminary instances, which combine the semantic prediction and the density of each point to group points in the offset coordinate system. Considering that cross-device variations can lead to changes in point cloud features and, thus, result in inaccurate predictions of point offset coordinates, we directly cluster the foreground points in the original coordinate system. The detailed clustering steps could be found in Alg.~\ref{alg}.

Furthermore, the MinkUnet~\cite{li2019gs3d} serves as the scoring network for assigning scores to each instance. The score for each instance represents the confidence of the instance segmentation, with higher scores indicating better instance segmentation. By analyzing the coordinates of each instance, the bounding box for each target can be computed. The center point coordinate $C \in \mathbb{R}^{N_{i n s} \times 3}$ of each object could be achieved by the following formula:
\begin{equation}
C_{i}=\frac{1}{N_{\operatorname{map}(i)}^I} \sum_{j \in I_{\operatorname{map}(i)}} p_j,
\end{equation}
where $\boldsymbol{p}_i=\left\{p_x^i, p_y^i, p_z^i\right\}$ describes the 3D coordinate of point $i$ in the original point cloud. Moreover, $N_{\operatorname{map }(i)}^I$ is the number of points in prediction instance $I_{\operatorname{map}(i)}$, and $N_{i n s}$ stands for the number of predicted instances.

\subsection{ScoreNet$^\ddagger$} 
Considering the significant gap between practical grasping scenarios and algorithmic simulations, we propose incorporating the number of points contained in the object proposal and the average height of the object proposal into the scoring system ScoreNet$^\ddagger$ to improve detection reliability and task suitability. This is because, on one hand, we observe that fragmentary point clouds are often misclassified as objects; on the other hand, top-down grasping conforms to a commonly applicable grasping strategy. To this end, we define the object detection score SC as follows:
\begin{equation}
S C= \begin{cases}-1 & N_{\text {map }}^I<N_\theta \\ \alpha \cdot S_f+(1-\alpha) H_m & N_{\text {map }}^I \geqslant N_\theta\end{cases},
\label{equ:score}
\end{equation}
where $S_f$ is the score of the backbone feature $F_b$, achieved by the ScoreNet. Moreover, $\alpha$ is the weight of the score $S_f$ within the range [0, 1]. And $N_\theta$ is a point number threshold; predictions with fewer points than this threshold are regarded as false positives, labeled as -1, and excluded from the grasp candidate set.

\subsection{Robot Grasping} 
The key to adopting 3D segmentation results for robotic object grasping lies in two aspects: the center point of the grasp and the grasp’s posture. The center point  $Cr \in \mathbb{R}^{N_{i n s} \times 3}$ of the grasp could be obtain by the instance center $C \in \mathbb{R}^{N_{i n s} \times 3}$:
\begin{equation}
Cr = C \cdot R + T, 
\label{eq:3}
\end{equation}
where $R$ and $T$ are the rotation matrix and translation matrix, respectively, obtained from the hand-eye calibration operation. In addition, 
we suspend the robot’s hand above the target object and calculate the yaw angle $\theta$ to grasp the target object. $\theta$  could be calculated by the point cloud coordinates:
\begin{equation}
\theta = arctan(\frac{y_{xmax} - y_{min}}{x_{ymax} - x_{min}}),
\label{eq:4}
\end{equation}
where $y_{min}$ and $x_{min}$ represents the minimum value of $x$ and $y$ coordinates in the point cloud. Additionally, $y_{xmax}$ denotes the value of $y$, when  $x$ achieves the maximum. Similarly, $x_{ymax}$ stands for the value of $x$, when $y$ attains the maximum. In our future work, we will consider the open pose~\cite{xia2025towards,zhao2025open} of the object and analyze the grasping pose from multiple perspectives.

\subsection{Pseudo-code} 
For a clearer understanding, we provide the pseudo-code of our cross-device robotic grasping pipeline in Alg.~\ref{alg}. 

\begin{algorithm}[h]
	\caption{Cross-device robotic grasping policy}
	{ \textbf{Data:} Object confidence threshold: $C_{\theta}$
    
         \qquad  \quad Point Density threshold: $d_{\theta}$}
        \begin{algorithmic}[1]
        
        \WHILE{True}
        \STATE Input the newest 3D point cloud scene
        \STATE  Obtain the foreground point $F_p$, according to the Eq.~\ref{eq:1} and Eq.~\ref{eq:2}
        \STATE Create the point set $H_{set}$ and $L_{set}$ 
        \IF{The density of $F^i_p >  d_{\theta}$ } 
        \STATE $H_{set}.push\_back(F^i_p)$ 
        \ELSE 
        \STATE $L_{set}.push\_back(F^i_p)$
        \ENDIF
        \STATE Obtain the initial instances $I$ by $Group(H_{set})$
        \STATE Refine the instances $I$ by $Voting(L_{set})$
        \STATE Attain the confidence scores $SC$ of instances $I$ by $ScoreNet^\ddagger(I)$
        \IF{$max(SC)<C_{\theta}$}
         \STATE break
         \ENDIF
        \STATE Confirm the highest confidence instance $I_{idx}$ by $idx= argmax(SC)$
        \STATE Grasp instance $I_{idx}$ by Eq.~\ref{eq:3} and Eq.~\ref{eq:4}
        \ENDWHILE
	\end{algorithmic}
\label{alg}
\end{algorithm}

\begin{figure*}[h]
  \begin{subfigure}{0.96\textwidth}
    \includegraphics[width=\linewidth]{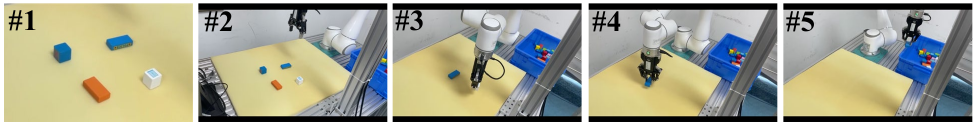}
    \caption{Scenario 1. \textbf{\#1} represents the initial grasping scenario and also the simplest one, where objects are not stacked and are sampled from the closed-set dataset. \textbf{\#2}, \textbf{\#3}, and \textbf{\#4} illustrate the grasping process of System~I, while \textbf{\#5} indicates all samples are successfully grasped.}
  \end{subfigure}
  
  \begin{subfigure}{0.96\textwidth}
    \includegraphics[width=\linewidth]{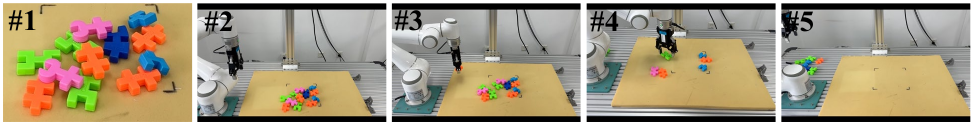}
    \caption{Scenario 2. \textbf{\#1} represents the initial grasping setup, in which all objects are from unseen categories and consist of various building blocks arranged in random stacks to increase scene complexity.  \textbf{\#3} and \textbf{\#4}  demonstrate that our policy possesses robust recognition and stable grasping capabilities for objects of various shapes, while \textbf{\#5} indicates all samples are successfully grasped.}
  \end{subfigure}
  
  \caption{Experiments with System I.}
  \label{fig:sys1}
\end{figure*}

\section{Experiment}

\subsection{Experiment Settings}
\textbf{Hardware} The two cross-device hardware configurations are as follows: System I is equipped with an Elite EC66 robotic arm and an AiNSTEC\_PRO\_L\_200\_5W 3D vision camera; System II integrates a Dobo CR robot with an Intel RealSense D455 vision camera. Both systems adopt an eye-to-hand configuration. The AiNSTEC\_PRO camera is set by default to a resolution of 1624×1240 with a depth accuracy of 0.1\%, while the RealSense D455 camera is configured by default to 1280×720 with a depth accuracy of 2\%.  Our code has been validated to operate on a single RTX 1080 Ti, RTX 2080 Ti, RTX 3090, and RTX 4090 GPU.

\noindent\textbf{Software} Our 3D object detection is implemented using Python, C++, and CUDA programming, while motion control of the robotic arm is based on the SDK provided by the manufacturer. All the code can be run on Windows using the WSL (Windows Subsystem for Linux) service. Following the existing segmentation works~\cite{jiang2020pointgroup,pbnet,wu2024point,zhao2025bfanet}, we adjust the voxel size to 0.005 for feature extraction. Moreover, parameters $N_\theta$ and $\alpha$ are set to 60 and 0.3, respectively. And parameters $C_\theta$ and $d_\theta$ are set to 0.5 and 2, respectively.

\noindent\textbf{Metrics} We evaluate our policy from two perspectives. First, based on instance segmentation results on benchmark datasets, we adopt commonly used metrics in the current literature~\cite{dai2017scannet,pbnet,yin2024sai3d}, including mean average precision $AP$ ($mAP$) at overlap 0.25 ($AP_{25}$), overlap 0.5 ($AP_{50}$), and over overlaps in the range [0.5:0.95:0.05] ($AP$). Second, we assess the detection and grasping accuracy in real-world open environments to validate the effectiveness and robustness of our approach under practical conditions.

\subsection{Clustering-based Methods Comparison}
We revisit 3D detection methods based on clustering to seek the best performance of our training-free grasp policy. The distance clustering in PointGroup~\cite{jiang2020pointgroup} is probably the most popular method, which has been adopted in Dyco3D~\cite{he2021dyco3d}, HAIS~\cite{chen2021hierarchical}, MaskGroup~\cite{zhong2022maskgroup}, SoftGroup~\cite{vu2022softgroup}, and RPGN~\cite{dong2022learning}. On the other hand, the binary clustering in PBNet~\cite{pbnet}  exhibits the highest performance on the public ScanNetv2 dataset, as recorded in Tab.~\ref{Tab:com_clustering}. To this end, we incorporate these two clustering methods into our reference pipeline to evaluate their performance in robot grasping on the Boxes dataset, respectively. 

\begin{figure}[h]
	\centering
	\includegraphics[width=0.47
    \textwidth]{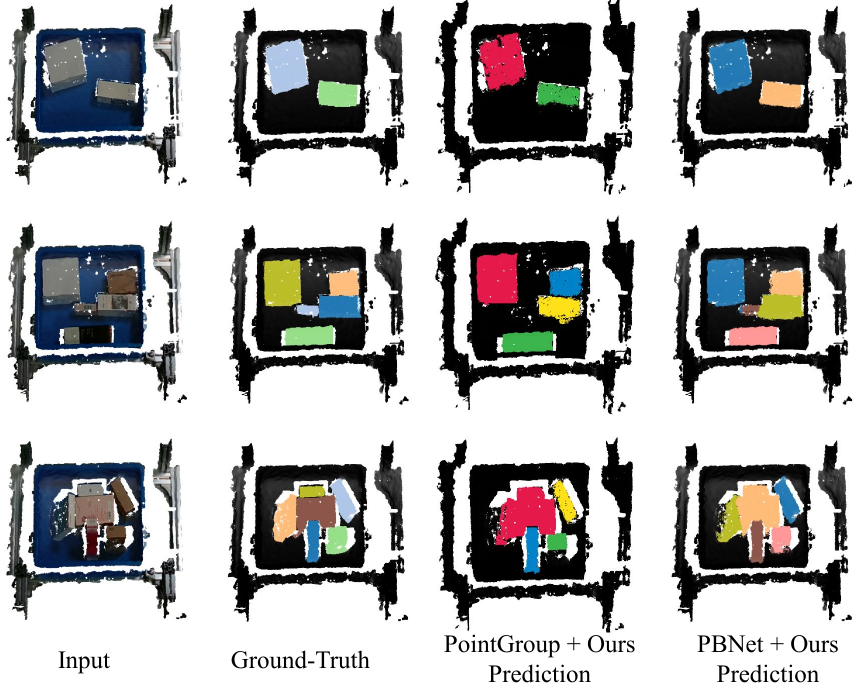} 
	\caption{Segmentation comparison on boxes dataset. Ours stands for our policy.}
	\label{fig:vis_com}
\end{figure}

Specifically, "PointGroup + Ours'' denotes the use of PointGroup as the closed-set 3D object detection method in Fig.~\ref{fig:pipeline}, while "PBNet + Ours'' represents the use of PBNet. As reported in Tab.~\ref{Tab:com_clustering}, our policy demonstrates strong adaptability to both methods, which indirectly reflects its high generalization capability.
Moreover, "PBNet + Ours'' is better than "PointGroup + Ours'' on these three metrics with a performance gain of  27.4\%, 12.0\%, and 8.0\%, respectively. Qualitatively, "PBNet + Ours'' also outperforms "PointGroup + Ours'', particularly in stacked scenes as shown in Fig.~\ref{fig:vis_com}. Building upon these observations, we choose PBNet as the closed-set 3D object detection method for our cross-device robotic grasping experiments in the real world.

\begin{table}[h]
\centering

\resizebox{0.47\textwidth}{!}{
\begin{tabular}{c|c|ccc}
\hline
Datasets                   & Methods      & $mAP$  & $AP_{50}$ & $AP_{25}$ \\ \hline
\multirow{7}{*}{ScanNetv2} & PointGroup   & 34.8 & 56.9 & 71.3 \\
                           & Dyco3D       & 40.6 & 61.0 & -    \\
                           & HAIS        & 43.5 & 64.1 & 75.6 \\
                           & MaskGroup    & 42.0& 63.3& 74.2 \\
                           & SoftGroup  & 46.0 & 67.7 & 78.9 \\
                           & RPGN         & -    & 64.2 & -    \\
                           & PBNet      & \textbf{54.3} & \textbf{70.5} & \textbf{78.9} \\ \hline
\multirow{2}{*}{Boxes}     & PointGroup + Ours& 51.6 & 84.0 & 88.0 \\
                           & PBNet + Ours    & \textbf{79.0} & \textbf{96.0} & \textbf{96.0} \\ \hline
\end{tabular}
}

\caption{Clustering-based methods  compared on 3D point cloud datasets. Ours stands for our policy.}
\label{Tab:com_clustering}
\end{table}

\begin{figure*}[h]
  \centering
    
  \begin{subfigure}{0.96\textwidth}
    \includegraphics[width=\linewidth]{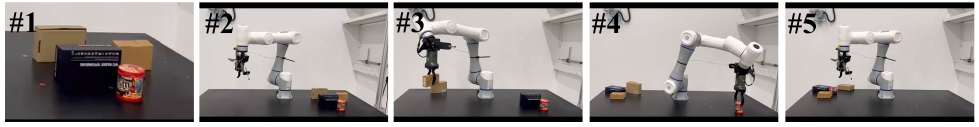}
    \caption{Scenario 3. \textbf{\#1} serves as the initial and relatively simple grasping setup, in which the target objects to be grasped are composed of common items from the real world. Clearly, these samples are entirely distinct from the seen categories used during pre-training. \textbf{\#2}, \textbf{\#3}, and \textbf{\#4} demonstrate that our policy is also capable of effectively recognizing and successfully grasping these samples. \textbf{\#5} indicates all samples are successfully grasped.}
  \end{subfigure}

  \begin{subfigure}{0.96\textwidth}
    \includegraphics[width=\linewidth]{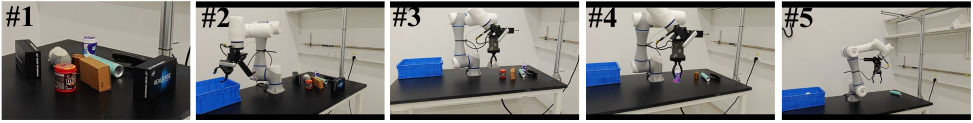}
    \caption{Scenario 4. In \textbf{\#1}, we introduce a greater variety of common objects, such as dolls and snacks, to further evaluate the generalization capability of our policy. \textbf{\#2}, \textbf{\#3}, and \textbf{\#4} illustrate the grasping process of System~II. In \textbf{\#5}, a long spray can is correctly identified but fails to be grasped, as it slips from the gripper due to its smooth surface.}
  \end{subfigure}

  \begin{subfigure}{0.96\textwidth}
    \includegraphics[width=\linewidth]{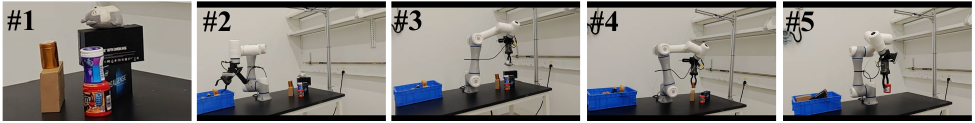}
    \caption{Scenario 5. In \textbf{\#1}, we increase the difficulty of the grasping scenario by stacking the objects. \textbf{\#3}, and \textbf{\#4} suggest that our policy prioritizes grasping objects at higher elevations, as our ScoreNet$^\ddagger$ incorporates object height as an additional criterion in the Eq.~\ref{equ:score}. \textbf{\#5} indicates all samples are successfully grasped.}
  \end{subfigure}
  
  \caption{Experiments with System II.}
  \label{fig:sys2}
\end{figure*}

\subsection{Cross-Device Robotic Grasping Comparison}
We conduct cross-device grasping experiments on two systems with entirely different hardware configurations: System I is equipped with an Elite EC66 robotic arm and an AiNSTEC PRO L 200 5W 3D vision camera; System II integrates a Dobo CR robot with an Intel RealSense D455 vision camera. As shown in Fig.~\ref{fig:sys1} and Fig.~\ref{fig:sys2}, we validate the system~I in two scenarios, and system~II in three scenarios. We also provide grasping videos in the supplementary material. Furthermore, we report the quantitative results in Tab.~\ref{tab:com_v}. Clearly, our policy can be effectively applied to open-category 3D object grasping across multiple different devices, without requiring any additional training or fine-tuning. Detailed qualitative discussions can be found in the captions beneath the corresponding figures. We summarize several additional key findings from our experiments as follows:

\noindent \textbf{Key Finding 1: The geometric structural relationship is vital.} Minimizing reliance on visual deep learning features and emphasizing the geometric properties of point clouds can effectively enhance robustness and generalization in cross-camera deployments.

\noindent \textbf{Key Finding 2: Cleaner data, more robust grasping.} Maintaining data consistency and high confidence enhances grasping robustness, through preprocessing, density normalization, and uniform scaling. Multi-perspective confidence evaluation of features and predictions further filters out low-quality data.

\setlength{\tabcolsep}{0.05cm}
\begin{table}[h]
    
    \centering
    \begin{tabular}{ccccccccc}
        \hline
        \noalign{\smallskip}
                                    & \multicolumn{4}{|c}{Scenario 1} & \multicolumn{4}{|c}{Scenario 2}                                                                                                                     \\
        \noalign{\smallskip}
        \hline

                                    & \multicolumn{1}{|c}{Recog.}     &                                 &                                 & \multicolumn{1}{c|}{Grasp.} & Recog.                      &        &   & Grasp. \\
        \hline
        \noalign{\smallskip}
        System I                      & \multicolumn{1}{|c}{100\%}      &                                 &                                 & 100\%                       & \multicolumn{1}{|c}{100\%}  &        &   & 100\%  \\
        \noalign{\smallskip}
        \hline
        \noalign{\smallskip}

        \noalign{\smallskip}
    \end{tabular}
    \begin{tabular}{ccccccccc}
        \hline
        \noalign{\smallskip} & \multicolumn{2}{|c}{Scenario 3} & \multicolumn{2}{|c}{Scenario 4} & \multicolumn{2}{|c}{Scenario 5}                                                                                                                                                                                    \\
        \noalign{\smallskip}
        \hline
                                    & \multicolumn{1}{|c}{Recog.}     & Grasp.                          & \multicolumn{1}{|c}{Recog.}     & Grasp.                      &
        \multicolumn{1}{|c}{Recog.} & Grasp.                                                                                                                                                                                \\
        \hline
        \noalign{\smallskip}
        System II                    & \multicolumn{1}{|c}{100\%}      & 100\%                           & \multicolumn{1}{|c}{100\%}      & 88.9\%                      &
        \multicolumn{1}{|c}{100\%}  & 100\%                                                                                                                                                                                 \\
        \noalign{\smallskip}
        \hline
    \end{tabular}
    
     \caption{
    Recognition rate (Recog.) and grasping rate (Grasp.) of our policy across various open-world scenarios with different device combinations. }
    \label{tab:com_v}
\end{table}

\section{Limitation and Future Work}

\textbf{Category-agnostic 3D object detection.} Our policy directly adapts pre-trained closed-set 3D object detection methods to an open-world arbitrary object grasping setting. Since open-world category information is absent in the pre-trained models, we disregard object category information entirely and proceed to grasp all detected objects based on their confidence scores. In future work, we plan to incorporate Vision-Language Models (VLMs) to generate pseudo-labels for open-world objects.

\noindent \textbf{Constrained grasp pose estimation.} Our work primarily focuses on cross-device  (cameras \& robotics) grasping and open-category grasping, for which we employ only a conventional method for grasp pose estimation. While more accurate grasp pose estimation approaches have been proposed in recent studies~\cite{hoang2024graspability,thalhammer2024challenges}, we plan to incorporate these advanced methods into our system in future work

\section{Conclusion}
In this work, we make the early attempt to design a policy to mitigate the impact of different devices (cameras \& robotics) for 3D open-world grasping. Specifically, we introduce the SSGC-Seg module to adapt closed-set 3D detectors for open-world object proposal detection, and develop ScoreNet$^\ddagger$ to enhance detection and grasping robustness. Notably, the proposed policy exhibits considerable adaptability, maintaining seamless compatibility with most existing clustering-based 3D object detection methods. Experiments conducted on segmentation datasets and in real-world grasping environments comprehensively validate the effectiveness of our policy. In future work, we plan to explore the integration of VLMs to assign pseudo-labels to open-world objects and to optimize our grasp pose estimation algorithm.

\bibliography{aaai2026}
\newpage
\end{document}